\documentclass[11pt,a4paper]{article}
\usepackage[hyperref]{acl2020}
\usepackage{times}
\usepackage{latexsym}
\usepackage{booktabs}
\usepackage{url}
\usepackage{tikz}
\usepackage{pgfplots}
\usepackage{fontawesome}
\usepackage{pifont}
\newcommand{\cmark}{\ding{51}}%
\newcommand{\xmark}{\ding{55} }%

\usepackage{amssymb}

\aclfinalcopy 

\title{Induced Inflection-Set Keyword Search in Speech}

\author{Oliver Adams,\textsuperscript{$\heartsuit\spadesuit$} Matthew Wiesner,\textsuperscript{$\spadesuit$} Jan Trmal,\textsuperscript{$\spadesuit$}\\
        {\bf Garrett Nicolai,\textsuperscript{$\diamondsuit\spadesuit$} David Yarowsky\textsuperscript{$\spadesuit$}} \\
  \textsuperscript{$\spadesuit$}Center for Language and Speech Processing, Johns Hopkins University, USA \\
  \textsuperscript{$\heartsuit$}Atos zData, USA \\
  \textsuperscript{$\diamondsuit$}Department of Linguistics, University of British Columbia, Canada\\
  {\tt oliver.adams@gmail.com,gnicolai@mail.ubc.ca},\\
  {\tt \{mwiesner,yenda,yarowsky\}@jhu.edu}}

\date{}

\begin{document}
\maketitle
\begin{abstract}

    We investigate the problem of searching for a lexeme-set in speech by searching for its inflectional variants. Experimental results indicate how lexeme-set search performance changes with the number of hypothesized inflections, while ablation experiments highlight the relative importance of different components in the lexeme-set search pipeline and the value of using curated inflectional paradigms. We provide a recipe and evaluation set for the community to use as an extrinsic measure of the performance of inflection generation approaches.


\end{abstract}

\section{Introduction}

Keyword search (KWS) is the task of finding certain words or expressions of interest in a body of speech. KWS is relevant to incident-response situations such as those modeled by LORELEI \cite{strassel2016lorelei} and was a focus of the IARPA Babel Program.\footnote{www.iarpa.gov/index.php/research-programs/babel} In the event of a humanitarian crisis, processing speech to determine mentions of certain keywords can inform better decision making when time is critical.

KWS is typically framed as searching for instances of a keyword in lattices that result from speech recognition decoding, as this means search is not restricted to a potentially incorrect one-best transcription. However, existing work on KWS assumes the relevant form of a keyword has been correctly specified. Many concepts to be searched for in speech take different forms through inflection as a result of the language's morphosyntax. In most cases, distinctions between such inflections (e.g.\ kill, kills, killing, killed) are irrelevant to the problem of searching for the underlying concept of interest.

Producing such inflection sets manually is arduous, even for native speakers,
 yet curators of keyword lists may have to construct them cross-lingually using bilingual dictionaries, which typically only contain canonical forms. Compounding this issue are the limitations of existing language technology for most of the world's languages across the whole KWS pipeline, including inflection generation,
 the language model (LM), the pronunciation lexicon, and the acoustic model. 

\begin{figure}
\input{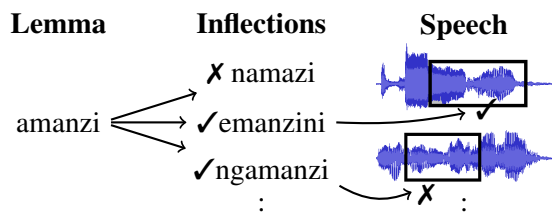}
\caption{An example Zulu keyword lemma (left) is inflected (middle) and then searched for in a corpus of speech (right). \cmark and \xmark  indicate correct/incorrect inflections, and correct/incorrect findings of the inflection in the corpus.}
\label{fig:illustration}
\end{figure}


In this paper we explore the application of inflection generation to KWS by searching for instances of a lexeme (see Figure \ref{fig:illustration}). To the best of our knowledge, this task has not been investigated before. Using Bengali and Turkish as evaluation languages, we scale the number of inflections generated per lexeme-set to examine how the trade-off between false positives and false negatives affects downstream KWS. We additionally perform experiments that assume varying quality of inflection generation.
Our findings show that lexeme-set KWS yields promising results even when all inflections must be generated on the basis of a distantly supervised cross-lingual approach to training inflection tools, though we observe that having a curated set of inflectional paradigms is important for achieving good performance. These first results encourage future work for lexeme-set search in speech, and the use of KWS as an extrinsic evaluation of inflection generation tools. 

To this end, we make available to the community a lexeme-set KWS pipeline with baseline models for inflection generation, grapheme-to-phoneme conversion (G2P), multilingual acoustic modeling, and cross-lingual inflection generation,
and a KWS evaluation set built on a suitable intersection of UniMorph inflection sets \cite{sylak-glassman2015language,kirov2018unimorph} and the Babel speech \cite{babel_turkish,babel_bengali}. The combination of these components serves as a novel downstream evaluation of inflection generation approaches, as well the other components in the pipeline. We make this recipe and evaluation set freely available online.\footnote{https://github.com/oadams/inflection-kws}




\section{The lexeme-set KWS Pipeline}

The pipeline starts with a lexeme of interest from the evaluation set (\S\ref{sec:eval_set}). Inflections of the lexeme are generated using some generation tool or manual resource (\S\ref{sec:inflection_generation}). These inflections are then converted to a phonemic representation (\S\ref{sec:g2p}) before being added to the lexicon used in speech recognition. KWS is then performed (\S\ref{sec:kws}) by decoding the speech and the model is scored on the how well it finds instanes of the lexeme.




\subsection{Evaluation Set}
\label{sec:eval_set}

We evaluate systems based on their ability to complete the following task: given a lemma, find all occurrences of its inflections in speech. To create an evaluation set for this task, we use UniMorph data, which provides ground truth inflection sets for a substantial number of languages. We use as our evaluation set instances of words in the Babel 10h development set that also are inflections in the UniMorph data. We remove from this set a small number of inflections that occur in more than one paradigm, as well as those that don't occur in the Babel pronunciation lexicon. This means that we can use the Babel lexicon as an oracle pronunciation lexicon with respect to our constructed evaluation sets to compare against our other methods. The result is an evaluation set tailored to morphologically salient word forms, with 1250 Turkish paradigms and 59 Bengali paradigms. The set of evaluation languages that can be extended to other languages in the Babel set for which we have ground truth paradigms.


\subsection{Inflection Generation}
\label{sec:inflection_generation}

Inflection generation is the task of producing
an \emph{inflection}, given a lemma and a
\emph{bundle} of morphosyntactic features.
For example, {\tt{run}} + \{\textsc{Pres;3;Sg}\} 
{$\mapsto$} ``runs''.
The state of the art in inflection generation
has arisen from the CoNLL--SIGMORPHON Shared 
Tasks~\cite{cotterell2016sigmorphon,cotterell2017conll,cotterell2018conll, mccarthy-etal-2019-sigmorphon}, and typically
consists of a modified sequence-to-sequence model with 
attention~\cite{makarov2018}.

However, these systems are fully supervised,
and hand-curated morphological dictionaries
often
do not exist.
We instead turn to the methods of \citet{nicolai-yarowsky-2019-learning}, who use English annotation as 
distant supervision to induce target language
morphology, using a widely-translated, 
verse-parallel text: the Bible.
Starting from the inflection pairs extracted by 
their method, we ensemble generators trained 
using an RNN and 
DirecTL+~\cite{jiampojamarn2010}.
For each lemma in the respective UniMorph,
we generate hypotheses for each feature
bundle, ensembling via a linear combination
of confidence scores. This gives us a set of inflections for each of the lexemes in the evaluation set which can then be searched for in the speech.



\subsection{Grapheme-to-Phoneme Conversion}
\label{sec:g2p}
To include hypothesized inflections in the KWS pipeline, orthographic forms of inflections must be mapped to a phonemic form consistent with the units used by the acoustic model \cite{maskey2004boostrapping,chen2016acoustic,mortensen2018epitran,schultz2007spice,KominekLearning,deri2016grapheme,trmal2017kaldi}. We use a finite-state transducer model trained with Phonetisaurus\footnote{\url{github.com/AdolfVonKleist/Phonetisaurus}} on 5,000 word forms in the target language.

\subsection{Keyword Search}
\label{sec:kws}

After generating inflections of lemmas in the evaluation set, these inflections are then included in the lexicon used in KWS. The KWS involves decoding the speech into lattices, and assessing lattice's inclusion of the keyword of interest. Our pipeline builds on the Kaldi OpenKWS system \cite{trmal2017kaldi}, which uses the standard lattice indexing approach of \cite{can2011lattice}. We use augmented pronunciation lexicons for KWS, which has been shown to outperform proxy KWS, a popular alternative \cite{chen2013using}.

The novel problem of lexeme-set KWS is related to work on out-of-vocabulary KWS, which has been approached by handling sub-word units such as syllables and morphemes \cite{trmal2014keyword,narasimhan2014,heerden2017,he2016}. In contrast to KWS with sub-word granularity, our approach is to generate likely full-word inflections given a lemma.






For language modeling, we used a 4-gram modified Kneser-Ney baseline \cite{kneser1995improved}. We compare using as training data the in-domain Babel text to the Bible, a resource available for many languages, and which was the resource used for cross-lingual distant supervision for inflection generation described in Section \ref{sec:inflection_generation}. Hypothesized inflections not seen in the training data receive some probability mass in language model smoothing as is default in SRILM \cite{stolcke2002srilm}, the language modeling tool used.

Though a monolingual acoustic model could have been used, we chose to use a ``universal'' phoneset acoustic model which can effectively be deployed on languages not seen in training and is motivated by work in multilingual acoustic modeling \cite{Schultz2001,le2005first,stolcke2006cross,vesely2012language,vu2012multilingual,heigold2013multilingual,scharenborg2017building,karafiat2018analysis}. We train an acoustic model on 300 hours
of data from 25 languages using a common
phonemic representation across languages. The training data includes ~10 hours
for each of 21 different languages from the IARPA Babel
corpus,  a  20 hour  subset  of  the  Wall  Street  Journal,\footnote{LDC94S13B}
Hub4  Spanish  Broadcast  news,\footnote{LDC98T29}  and  the  Russian
and French portions of the Voxforge\footnote{http://voxforge.org} corpus.

\section{KWS Evaluation Metrics}

We evaluate KWS performance on a per lexeme-set basis, rewarding the system when it finds any form of an evaluation lexeme, regardless of how it is inflected, while also penalizing failure to find any inflection.

As an evaluation metric we use term weighted value (TWV), a standard metric in KWS developed for the NIST~2006 Spoken Term Detection evaluation \cite{fiscus2007results}, which rewards joint maximization of recall with minimization of false positives. TWV relies on a threshold parameter to determine what minimum level of confidence is required by the system in order to assert keyword findings. There are several variations of term weighted value (TWV) that are different in the way the threshold is handled: Actual (ATWV), Optimum (OTWV), and Supreme (STWV).

ATWV is the TWV of the system given some global threshold (provided by the system) of confidence common to all keywords, and is the most common metric used to compare systems.


OTWV determines a per-keyword (in our case, per lexeme-set) threshold. For our purposes this is the most informative metric because it gives a better sense of how the ATWV would be if system effectively normalized confidences across lexemes. Improvements to TWV may also potentially be made beyond what is represented by the OTWV. Some inflections are more likely than others, yet the thresholds for OTWV are made at a per lexeme-set basis, not a per-inflection basis. Improving how the system weights the likelihood of different inflections (either during inflection generation or in the LM probabilities) would likely substantially improve ATWV.

STWV is a recall-oriented version of TWV that disregards the confidence of the terms and does not penalize false positives. It is thus similar to lattice recall and serves as a useful metric in system analysis for determining whether low ATWV/OTWV is due to large number of false positives or issues in effective speech word lattice decoding.





\section{Experiments}

We conduct experiments to see how performance of KWS relates to the number of inflections hypothesized by the cross-lingual distantly-supervised method described in Section \ref{sec:inflection_generation} (henceforth \texttt{RNN+DTL}), before comparing it to several alternative benchmark methods.

As evaluation languages we used Bengali and Turkish, a subset of languages for which we have Bibles and that also occur in UniMorph. We observed similar trends and relative performance of methods for both languages languages so in the subsequent results we present the arithmetic mean of the results of Bengali and Turkish.

\subsection{The Number of Generated Inflections}
\label{sec:num_inflections}

\begin{figure}
    \centering
    \begin{tikzpicture}
\begin{axis}[
		width=7cm,
		height=5cm,
		yticklabel style={
				/pgf/number format/fixed,
				/pgf/number format/precision=1,
				/pgf/number format/fixed zerofill
		},
		ylabel={TWV},
		xlabel={$k$ (\# generated inflections/bundle)},
		legend columns=-1,
		legend entries={ATWV;, OTWV;, STWV},
		legend style={at={(0.5,-0.4)},anchor=north},
		ymin=-0.20,
		ymax=1.0,
		xmode=log,                                                              
        log ticks with fixed point,
        ytick={0.0,0.20,0.4,0.6,0.8,1.0}
		]
\addplot[color=black,dashed, mark=x, mark options={solid}]
	coordinates {
(1, 0.025)
(5, 0.043)
(10, 0.110)
(20, 0.1229)
(40, 0.133)
(80,-0.107)
(160,-0.295)
	};
\addplot[color=black,solid, mark=x, mark options={solid}]
	coordinates {
(1, 0.0323)
(5,0.064)
(10,0.153)
(20,0.2328)
(40,0.269)
(80,0.212)
(160,0.109)
	};
\addplot[color=black,dotted, mark=x, mark options={solid}]
	coordinates {
(1,0.04225)
(5,0.109)
(10,0.235)
(20,0.45725)
(40,0.577)
(80,0.719)
(160,0.764)
	};
\end{axis}
\end{tikzpicture}\\
    \vspace{-1.5em}
    \caption{Variations on the term weighted value (TWV) metric for different numbers of generated inflections per morphosyntactic bundle (by \texttt{RNN+DTL}). Overgeneration of inflections improves recall, as captured by STWV, but leads to too many false positives when $k\geq80$.}
    \label{fig:curve}
\end{figure}
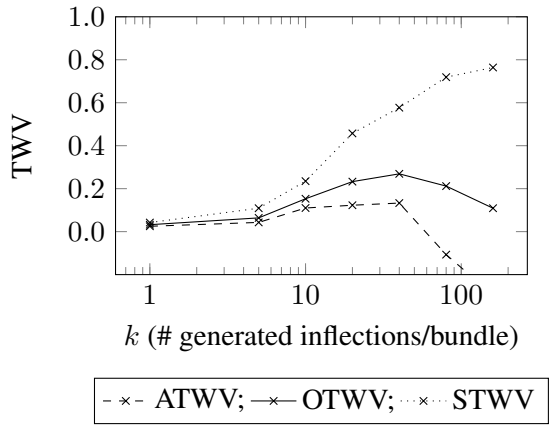

To gauge how over-generation of inflections affects KWS performance we scaled $k$, the number of inflections generated per morphosyntactic bundle. Figure \ref{fig:curve} illustrates how TWV varies with respect to $k$ for inflections generated by \texttt{RNN+DTL} in a KWS system that uses the in-domain Babel LM. At values of $k$ beyond 40 the ATWV and OTWV began to decrease, as the number of false positives was too high. The recall-oriented STWV continued to increase, peaking at 0.764 when $k=160$. It is noteworthy that OTWV only began to decrease at such a value of $k$. For Turkish nouns, with 23 bundles per lexeme-set, a value of $k=40$ corresponds to 920 inflections, the vast majority of which are invalid inflections. This indicates that there is room for a substantial amount of inflection overgeneration in KWS, since the speech recognition can provide acoustic evidence against incorrect inflection candidates.


\subsection{Comparison of Inflection Approaches}

To get a comparative sense of the KWS performance of \texttt{RNN+DTL} at the best value of $k$, we compare it with three other approaches: \texttt{Oracle}, \texttt{UniMorph}, and \texttt{Lemmas}, as shown in Table \ref{tab:ablation}.

\texttt{Oracle} includes exactly the set of inflections that occur in the evaluation set. \texttt{UniMorph} includes all the inflections that occur in the UniMorph data, which differs from \texttt{Oracle} in that it contains true inflections that don't happen to occur in the Babel speech. We included this to assess how true inflections of the lexeme that are not found in the speech affect performance. It helps substantially for ATWV and OTWV, but not for STWV. This somewhat counterintuitive result suggests that including more inflectional variants of a lexeme may not improve recall (i.e. improve STWV) but can decrease the number of false positives.

\begin{table}
\center
\resizebox{\columnwidth}{!}{%
\begin{tabular}{lccccc}
\toprule
Inflections & LM & ATWV & OTWV & STWV \\
\midrule
\texttt{Oracle} & Babel & 0.315 & 0.463 & 0.866\\
\texttt{UniMorph} &  Babel & 0.392 &	0.513 &	0.864 \\
\texttt{RNN+DTL} & Babel & 0.133 &	0.269 &	0.577 \\
\texttt{Lemmas} & Babel & 0.169 &	0.219 & 0.281\\
\midrule
\texttt{RNN+DTL-NS} & Babel & 0.304 & 0.443 &	0.815 \\
\texttt{RNN+DTL} & Bible & 0.046 &	0.206 &	0.561 \\
\bottomrule
\end{tabular}
}
\caption{Term weighted value (TWV) under varying conditions: \texttt{Oracle} inflections known to occur in the Babel speech; \texttt{UniMorph} inflections that additionally include true inflections not seen in speech; \texttt{RNN+DTL}-generated inflections via distant cross-lingual supervision; \texttt{Lemmas}-only search. Discounting spurious forms from \texttt{RNN+DTL} shows its high recall. Using an out-of-domain LM substantially decreases performance.
}
\label{tab:ablation}
\end{table}

\texttt{Lemmas} searches only for citation-form lemmas. It has a relatively decent ATWV (even outperforming \texttt{RNN+DTL}, though not by OTWV) despite low recall (as indicated by STWV) because it has few false positives and also because most inflections sound similar to the lemmas via the addition of an affix. As a result, searching for the lemma often catches inflectional variants too. 

We consider two further points of comparison. Firstly, \texttt{RNN+DTL} with only a Bible-trained LM, which underperforms other systems substantially except in lattice recall as indicated by STWV. Secondly, \texttt{RNN+DTL-NS}, which removes from \texttt{RNN+DTL} spurious inflections that weren't found in \texttt{Oracle}. Comparison of \texttt{RNN+DTL-NS} and \texttt{RNN+DTL} demonstrates that while the system has some robustness to overgeneration of inflected forms (\S\ref{sec:num_inflections}), it is also the case spurious inflections not only increase false positives, but can actually hurt recall too.

These results indicate that correctly generating inflected forms and properly weighting the hypothesized inflections (either via the inflection generation module, or in the language model) is the most critical bottleneck in the pipeline. The high relative performance of \texttt{Unimorph} indicates the value of making full use of available curated resources. Recent work has shown how effective inflection generation can be with limited resources in the target language \cite{cotterell2018conll}. These results suggest that if such resources are not available, then in practice it is likely worth gathering training data with which train an inflection generator.

\section{Conclusion}

We have presented an evaluation of lexeme-set KWS. Our results shed light on the relative impact of undergenerating and overgenerating inflected forms on KWS, indicating that high recall can be achieved via an inflection method of cross-lingual distant supervision, but with the best all-round performance achieved by making use of Unimorph.
We release our evaluation set along with scripts to reuse our pipeline so that the community can explore lexeme-set KWS as an extrinsic evaluation of inflection generation.

\section*{Acknowledgements}

We would like to thank all reviewers for their constructive feedback.

\bibliography{emnlp-ijcnlp-2019}
\bibliographystyle{acl_natbib}

\end{document}